\documentclass[twoside,11pt]{article}

\usepackage[accepted]{melba}

\usepackage{microtype}
\usepackage{subfigure}
\usepackage{hyperref} 
\usepackage{algorithmic}
\usepackage{graphicx}
\usepackage{textcomp}
\usepackage{tabularx}
\usepackage{array}
\usepackage{amsmath,amssymb,amsfonts}
\usepackage{booktabs}
\usepackage{dcolumn}
\usepackage{threeparttable}
\usepackage{multirow}
\usepackage{cleveref}

\newcommand\nw[1]{\textcolor{black}{#1}}
\newcommand\nwf[1]{\textcolor{black}{#1}}

\melbaid{2024:007}  
\doi{https://doi.org/10.59275/j.melba.2024-4dg2}
\melbaauthors{Gianchandani et al.}  
\volume{2}
\firstpageno{761}
\melbayear{2024}  
\datesubmitted{10/2023}  
\datepublished{04/2024}  

\melbaspecialissue{Medical Imaging with Deep Learning (MIDL) 2020}
\melbaspecialissueeditors{Marleen de Bruijne, Tal Arbel, Ismail Ben Ayed, Hervé Lombaert}

\ShortHeadings{Voxel-level brain age prediction to assess regional brain aging}{Gianchandani et. al.}

\title{A voxel-level approach to brain age prediction: \\A method to assess regional brain aging}

\author{\firstname Neha \surname Gianchandani \email neha.gianchandani@ucalgary.ca \\  
	\addr Department of Biomedical Engineering, University of Calgary, Canada\\
         Hotchkiss Brain Institute, University of Calgary, Canada \orcid{0000-0003-0822-4554}
	\AND
	\name Mahsa Dibaji \email seyedemahsa.dibaji@ucalgary.ca \\
	\addr Department of Electrical and Software Engineering, University of Calgary, Canada \orcid{0009-0004-3166-7737}
    \AND
    \name Johanna Ospel \email johanna.ospel@ucalgary.ca \\
    \addr Department of Radiology; Clinical Neurosciences, University of Calgary, Canada \orcid{0000-0003-0029-6764}
    \AND
    \name Fernando Vega \email fernando.vega1@ucalgary.ca \\
    \addr Department of Biomedical Engineering, University of Calgary, Canada \orcid{0000-0003-0013-8133}
    \AND
    \name Mariana Bento \email mariana.pinheirobent@ucalgary.ca \\
    \addr Department of Biomedical Engineering; Electrical and Software Engineering, University of Calgary, Canada\\
         Hotchkiss Brain Institute, University of Calgary, Canada \orcid{0000-0001-5125-0294}
    \AND
    \name M. Ethan MacDonald \email ethan.macdonald@ucalgary.ca \\
    \addr Department of Biomedical Engineering; Electrical and Software Engineering; Radiology, University of Calgary, Canada\\
         Hotchkiss Brain Institute, University of Calgary, Canada \orcid{0000-0001-5421-3536}
    \AND
    \name Roberto Souza \email roberto.souza2@ucalgary.ca \\
    \addr Department of Electrical and Software Engineering, University of Calgary, Canada\\
         Hotchkiss Brain Institute, University of Calgary, Canada \orcid{0000-0001-7824-5217}
}

\begin{document}

\maketitle

\begin{abstract}
	Brain aging is a regional phenomenon, a facet that remains relatively under-explored within the realm of brain age prediction research using machine learning methods. Voxel-level predictions can provide localized brain age estimates that can provide granular insights into the regional aging processes. This is essential to understand the differences in aging trajectories in healthy versus diseased subjects. In this work, a deep learning-based multitask model is proposed for voxel-level brain age prediction from T1-weighted magnetic resonance images. The proposed model outperforms the models existing in the literature and yields valuable clinical insights when applied to both healthy and diseased populations. Regional analysis is performed on the voxel-level brain age predictions to understand aging trajectories of known anatomical regions in the brain and show that there exist disparities in regional aging trajectories of healthy subjects compared to ones with underlying neurological disorders such as Dementia and more specifically, Alzheimer's disease.
	Our code is available at~\url{https://github.com/nehagianchandani/Voxel-level-brain-age-prediction}.
\end{abstract}


\section{Introduction}
\label{sec:introduction}
As humans progress through life and age, the brain ages as well and it can be observed with neuroimaging \citep{macdonald2021mri}. This concept, known as brain age, mirrors the chronological age but pertains specifically to the brain. It provides insights into the maturity level and developmental trajectory of an individual's brain which can sometimes be different from the overall aging process of an individual. For brain age studies, it is assumed that for healthy subjects, brain age is representative of chronological age, indicating that the brain is aging at a similar rate as humans age. However, for subjects with underlying neurological disorders, there is often a deviation in the aging trajectory. An effective biomarker of neurological disorders is increased brain age \citep{cole2017predicting,cole2018brain,huang2017age}. 

Early works on brain age provide a global estimate, \textit{i.e.}, brain age is studied as a single global index for the entire brain. Global brain age has been demonstrated as an effective biomarker to study the brain aging process in the presence and absence of various neurological disorders \citep{cole2017neuroimaging,franke2019ten}. However, due to its global nature, it does not provide spatial information on the brain aging process. Studies have shown that the aging process occurs at different rates and may be non-linear across different regions of the brain, highlighting region-specific response to the aging process \citep{hof1996neuropathological,raz2010trajectories}. The global brain age index is not able to capture this regional information related to aging. The concept of voxel-level brain age can help bridge the gap, where a voxel represents a small unit of the brain volume. Brain age prediction at the level of each voxel can provide a fine-grained analysis of how different regions of the brain age in healthy compared to diseased brains assigning a distinct brain age to each voxel of the brain. Voxel-level predictions can be particularly useful for understanding how neurological disorders impact different regions of the brain. Most neurological disorders are often associated with specific regions of the brain, for example, Alzheimer's disease (AD) is associated with atrophy in the hippocampus and temporal regions of the brain \citep{rao2022hippocampus,pasquini2019medial}, and Parkinson's is associated with basal ganglia \citep{blandini2000functional,caligiore2016parkinson}, and hence, these regions are expected to have an increased brain age as compared to other regions of the brain in the presence of corresponding disorders.

\begin{figure*}[h]
\centering
\includegraphics[width=\textwidth]{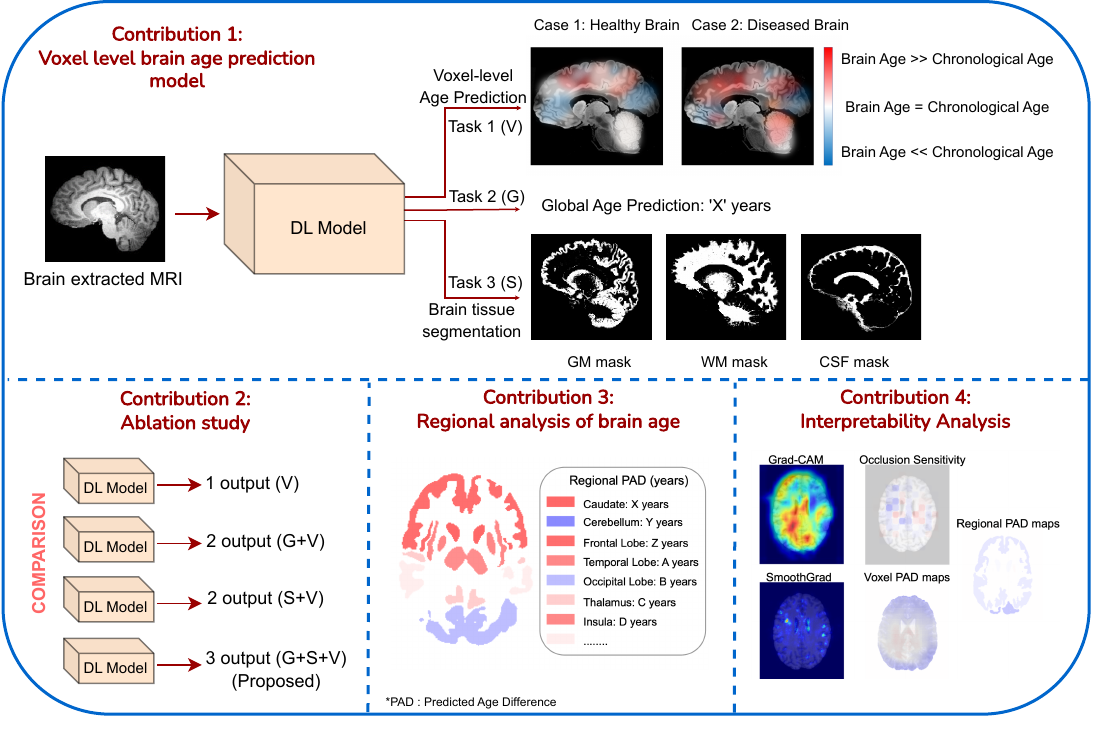}
\caption{Overview of the contributions of this article. There are 4 major contributions: (i) Proposal of a voxel-level brain age prediction model with initial validation on healthy subjects and subjects with dementia and Alzheimer's disease (ii) Ablation experiments were done to justify the use of a multitask architecture of the DL model with three-tasks over its two-task, and one-task counterparts. (iii) Regional analysis of predicted brain age by clustering voxel-level brain age predictions into known anatomical regions of the brain and (iv) An interpretability analysis where the proposed voxel-level approach to understanding regional aging trajectories is compared to traditional interpretability methods like Grad-CAM \nw{\citep{selvaraju2017grad}}, SmoothGrad \nw{\citep{smilkov2017smoothgrad}}, and Occlusion Sensitivity \nw{\citep{zeiler2014visualizing}}. }
\label{graphical_abstract}
\end{figure*}

In this article, an extended analysis and evaluation of our recently proposed deep learning (DL) model to predict voxel-level brain age using T1-weighted magnetic resonance (MR) images \citep{gianchandani2023} is presented. The initial work introduced a multitask architecture for voxel-level brain age prediction and evaluation of that model on presumed healthy subjects. In this work, the analysis is extended by performing an ablation study to reflect on how the multi-task architecture is an improvement over a single-task deep learning model. Additionally, the results of the proposed model are inspected and evaluated on subjects with dementia and more specifically, AD and report varying brain ages for different anatomical regions of the brain. A voxel-level brain age prediction model can provide an enhanced understanding of the regional aging processes in the brain while allowing the quantification of the deviation observed in years. Incorporating a multi-task framework moves closer to enhancing the transparency and interpretability of the DL model and it is substantiated by a comparison of the proposed methodology to existing interpretability methods implemented over a state-of-the-art global age prediction model. To summarize, the contributions of this paper are (refer to Figure \ref{graphical_abstract}):

\noindent 1. Proposal of a multitask DL voxel-level brain age prediction model, building upon our prior work \citep{gianchandani2023}, with an extended evaluation encompassing subjects with dementia and Alzheimer's disease.\\
2. An ablation study to show the importance of the different tasks in the multitask architecture.\\
3. Regional analysis of the brain aging process in presumed healthy subjects and subjects with dementia and specifically Alzheimer's disease. \\
4. Comparison of the proposed model with existing interpretability methods implemented over a state-of-the-art global age prediction model.

\section{Related Work}
\label{sec:litreview}
Brain age prediction is a well-researched domain, however, most studies focus on a global analysis of brain age. Initially, this was done with handcrafted features using traditional machine learning (ML) techniques like \nw{Support} Vector Machines, Random Forest, and other traditional machine learning models  \citep{valizadeh2017age, lemaitre2012normal,beheshti2021predicting}. The approach with traditional ML models is generally considered easier to explain and interpret owing to the reliance on simpler algorithms, fewer parameters, engineered features, and in-built feature importance scores, and achieved brain age predictions with mean absolute error (MAE) $\sim$ 4-8 years. The use of manually-engineered features can aid in understanding the model, but can also be restrictive at the same time as it can lead to the omission of crucial features during the feature engineering process. This limitation led the shift towards the use of DL models for predicting brain age. Manual feature engineering can inadvertently simplify and distort complex data representations, leaving scope for future improvements. Therefore, the transition to neural network models allowed to capture complex data representations within the data that are integral to this brain age prediction task \citep{plis2014deep}. DL models showed significant improvement in the brain age prediction task with MAE as low as 2-4 years \citep{ito2018performance,kolbeinsson2020accelerated}, however, due to the neural networks complexity, and black-box nature, these DL models have limited interpretability. 

Studies have attempted to explain DL models for brain age prediction with techniques like Grad-CAM \citep{bermudez2019anatomical}, saliency map-based techniques \citep{yin2023anatomically}, occlusion-map based techniques \citep{bintsi2021voxel}, layer-wise relevance propagation \citep{hofmann2022towards} and SHapley Additive ex-Planations (SHAP) \citep{ball2021individual}, among others, to better understand the regional contribution to the brain age prediction models. However, one common limitation of using existing interpretability techniques lies within the use of gradients to calculate feature importance and consequently, the inability to compare the relevance scores across samples. The explanations provided by the existing interpretability methods are quantitative, but only at a sample level as the relevance scores are based on the relative importance of different regions in the input image. Despite the flaws, the aforementioned methods have proven to be tremendously helpful in making the black-box models more transparent and a step closer to understanding the decision-making process of complex neural network architectures. Achieving state-of-the-art results should not come at the cost of interpretability. The proposed approach to predicting voxel-level brain age produces brain predicted age difference (PAD) maps that reflect on the regional aging processes and provide us with a way to quantify healthy versus diseased aging patterns of the brain that is comparable across samples. Additionally, the proposed modeling method ensures that structural features in the brain are used to predict brain age, this will be discussed in detail in Sections \ref{sec:results} and \ref{sec:discussion}.

To move towards a regional analysis of the brain aging process, studies \citep{beheshti2019novel,bintsi2020patch} have attempted to predict brain age at a block or a patch level (with an MAE in the range of $\sim$ 1.5-2 years) where predictions are made for individual blocks of the brain. These blocks do not necessarily correlate to known anatomical regions of the brain but do provide an additional  level of spatial information compared to the global-age prediction models. \nw{The authors suggest taking this a step further with an analysis at a higher resolution in future works.} It is important to acknowledge that studies have attempted to explore and understand regional aging trajectories in the brain using other techniques like regional volume changes \citep{raz2005regional}, functional changes \citep{davidson1999regional} etc., however, for the scope of this article, we will be limiting our focus on studies that utilize ML/DL techniques \nw{on anatomical images} to do so from a brain age prediction perspective. Finally, based on the current literature, voxel-level predictions have only been explored once before by \citet{popescu2021local}. Their method produces voxel-level age maps to understand the regional aging process in the brain, however, this is at the cost of a high MAE $\sim$ 9 years. The authors utilize a modified version of a U-Net architecture to predict brain age at a voxel-level and block-level. This method will be referred to as the baseline for the scope of this article. \nw{One common trend observed in most works on brain age prediction (global or regional) is the use of MAE being used as a metric for result comparison. MAE suffers from being influenced by the age range of the test set and hence, makes cross-study comparison less accurate. To overcome this limitation, we do report MAE like previous works, however, also report $R^2$ (Coefficient of Determination) and show violin plots for the results for a more holistic assessment of results.} 

\nw{Diverging from the baseline \citep{popescu2021local}, the proposed methodology introduces two significant modifications. First, the proposed method models brain age in the native space of the T1-weighted MR images, eliminating registration of any degree as a pre-processing step. This is done to ensure the features are retained in their truest form in the input images. Second, the proposed method uses full T1-weighted volumes as input to the brain age prediction model rather than segmentations of gray matter (GM) and white matter (WM) as done in the baseline. This ensures the inclusion of cerebrospinal fluid (CSF) in the input images which based on previous studies is relevant to the study of brain aging \citep{houston2023aging,may1990cerebrospinal}. The two improvements proposed in this manuscript are further examined in the discussion section.}

\section{Materials and Methods}
\label{sec:materialsandmethods}
\subsection{Data}
\label{subsec:data}
\nw{T1-weighted MR imaging is the most widely used MR sequence for brain age prediction \citep{cole2017predicting2,sajedi2019age}, likely due to the wide availability of T1-weighted data across a broad age range. Following the same, }T1-weighted MR imaging was utilized from publicly available datasets to encourage reproducibility of our work. All data corresponds to presumed healthy controls from the Cambridge Centre for Ageing Neuroscience (Cam-CAN) \citep{taylor2017cambridge} for training the model. \nw{The data was acquired on a Siemens TIM TRIO 3 T scanner.} The dataset (n=651) is nearly uniformly distributed across the age range of 18-88 years with a mean age of 54.24$\pm$18.56 years. The dataset has a sex-balance of 55\%:45\%, male:female ratio to limit sex-related bias in the model. 

An independent test set (n=359) corresponding to healthy controls for further validation of the model was sourced from the Calgary-Campinas-359 (CC359) dataset \citep{souza2018open} (age range 36-69 years with a mean of 53.46$\pm$9.72 years) with a balanced sex-distribution of 49\%:51\%. The CC359 dataset contains data acquired on scanners from three different vendors (Philips, General Electric [GE], Siemens) and at two different magnetic field strengths (1.5T, and 3T) giving rise to 6 subsets within the dataset to assess the robustness of the proposed model across different data acquisition protocols. 

To create the bias correction methodology (further discussed in Section \ref{subsec:biascorrection}), 48 healthy control subjects each from the Open Access Series of Imaging Studies (OASIS) \citep{marcus2007open}, Alzheimer's Disease Neuroimaging Initiative (ADNI) \citep{mueller2005alzheimer,mueller2005ways}, and Cam-CAN datasets (unseen during training) were extracted, totalling 144 samples. The ADNI was launched in 2003 as a public-private partnership, led by Principal Investigator Michael W. Weiner, MD. The mean age of the bias correction data set was \nw{$62.61\pm21.99$} years with a male:female ratio of \nw{50\%:50\%}. 

For the evaluation of the proposed model on subjects with underlying neurological disorders, two open-source datasets were utilized. Twenty-eight dementia subjects were extracted from the OASIS dataset \citep{lamontagne2019oasis} (mean age $69.17\pm5.13$ years) and twenty subjects with AD from the ADNI dataset  (mean age $64.8\pm5.24$ years). \nw{The data obtained from the OASIS dataset includes samples with varying dementia types and was chosen to include the analysis of subjects from the broad perspective of cognitive decline. For a more specific analysis, the ADNI dataset was chosen for the analysis of a specific kind of dementia, AD. The OASIS dataset was collected over a period of 10+ years on three different Siemens scanners, (i) Siemens Medical Solutions USA, Inc: Vision 1.5T, (ii) TIM Trio
3T (2 different scanners of this model), and (iii) BioGraph mMR PET-MR 3T. The ADNI samples were extracted from the ADNI 1 cohort which was acquired partly on 1.5T scanners using T1- and dual echo T2-weighted sequences and partly using the same protocol on 3T scanners. Detailed acquisition parameters for the datasets used are described in the source publication for each dataset.}

\subsection{Data preparation and pre-processing}
\label{subsec:dataprep}
To ensure that all MR images have the same orientation, FMRIB Software Library's (FSL) \citep{jenkinson2012fsl} `fslreorient2std' command was used. \nw{It is important to note that `fslreorient2std' is not a registration command, and hence, no registration is performed, rather it only applies 90, 180, or 270-degree rotations about the different axes as necessary to get the labels of the MR image in the same position as the standard MNI-152 template.} Brain extraction masks and tissue segmentation masks to segment GM, WM and CSF for the T1-weighted images from the Cam-CAN dataset were obtained using two U-Net models trained for the specific tasks on the CC359 dataset. The models were trained on the CC359 dataset due to the availability of the binary brain extraction masks and the tissue segmentation masks along with the publicly available T1-weighted images. The binary brain extraction masks are used to obtain brain-extracted input to the model and the tissue segmentation masks are used as ground truths for one of the output tasks in the methodology. All training MR images have a voxel size of 1 mm\textsuperscript{3}. \nw{Before feeding the MR images to the DL model, random rotation was implemented as an augmentation step. Patches of size $128\times128\times128$ were cropped from the MR images (1 crop per input image) to be used as input. Random cropping was performed such that the majority of samples have significant brain regions to ensure that the model learns relevant features. The models were implemented using PyTorch with MONAI \citep{cardoso2022monai} for the data preprocessing pipeline.}

\subsection{Proposed model architecture}
\label{subsec:proposedmodel}
\begin{figure*}[h]
\centering
\includegraphics[width=\textwidth]{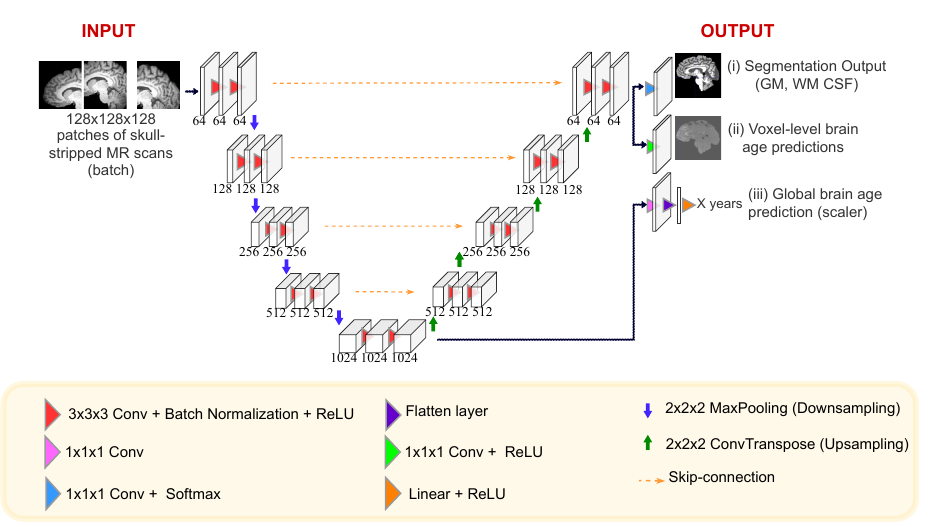}
\caption{\nw{The proposed multitask model used for voxel-level brain age prediction. The model follows a U-Net backbone with a downsampling and upsampling block to obtain the output at the same size as the input. It takes cropped MR patches of size $128\times128\times128$ as input to the model and produces three outputs, a voxel-level brain age prediction \textit{i.e.} the main output as well as two secondary outputs, brain tissue segmentation of GM, WM and CSF, and global brain age prediction.}}
\label{fig:proposed_model}
\end{figure*}

In this work, \nw{an extended evaluation of our recently} proposed multitask U-Net architecture \nw{\citep{gianchandani2023}} is presented to predict voxel-level brain age along with two additional tasks, global brain age prediction and brain tissue segmentation to segment GM, WM, and CSF. A multitask architecture refers to the presence of multiple outputs that the model is trained for simultaneously. Multi-task learning is known to improve the model training process by including multiple tasks for the model to learn shared representations on, this also helps in avoiding overfitting and leads to fast convergence \citep{crawshaw2020multi}. In the proposed methodology, the main task is the voxel-level brain age prediction task, to complement this task, a brain tissue segmentation task to segment GM, WM, and CSF and a global brain age prediction task are included. Global brain age prediction can be considered a simpler version of the voxel-level brain age prediction task. The segmentation task ensures that relevant structural features are learned from the MR data during training. \nw{A good segmentation performance will ensure the learning of structural features like GM, WM, and CSF thickness, shape changes, etc., and owing to the multitask nature of the model, the same features will be repurposed for the brain age prediction task. This ensures the reliance of brain age prediction on structural features in the brain volume.} The backbone of the proposed model is a simple U-Net architecture \citep{ronneberger2015u} that has an encoder and a decoder network, making a U-like shape. Batch-normalization layers are added after the convolution operations to ensure a smooth training process \citep{santurkar2018does}. The encoder and decoder are connected by skip connections that help with recovering important spatial information that is lost during downsampling. The model architecture is depicted \nw{in Figure \ref{fig:proposed_model}.}

\subsection{Loss function}
\label{subsec:loss}
To accommodate the multitask modeling approach with three different outputs, a custom loss function is defined to ensure that all tasks are given significant importance as the training progresses. The loss function for the proposed model is made up of three terms. $\text{Dice}_{\text{loss}}$ is used to accommodate the segmentation task and is computed from the Dice coefficient (DICE) based on Eq. \ref{eq:dice}. The Dice coefficient is a measure of the overlap between the ground truth $Y$ and predicted segmentation $\hat{Y}$. The DICE and $\text{Dice}_{\text{loss}}$ are inversely related, making the model learn accurate segmentations as $\text{Dice}_{\text{loss}}$ is minimized during the training process. \nw{$\text{Dice}_{\text{loss}}$ is first computed for all classes \textit{i.e.} background, GM, WM, and CSF individually and then averaged to obtain the the overall $\text{Dice}_{\text{loss}}$.}

\begin{equation}
\label{eq:dice}
\text{Dice}_{\text{loss}} = 1 - \text{Dice} =  1-{\frac{1}{m}\sum_{i=1}^m\frac {2|Y \cap \hat{Y}|}{|Y| + |\hat{Y}|}}
\end{equation}

MAE is the most commonly used metric for the loss function in brain age prediction studies \citep{feng2020estimating,bermudez2019anatomical,he2021global,popescu2021local}. The remaining two terms are two versions of MAE to accommodate the age prediction at the global and voxel-level.  Eq. \ref{eq:voxel} is the voxel-level MAE. First averaged across all brain voxels in the input, followed by batch average, where $y_{i,j}$ is the voxel-level brain age and $\hat{y}_{i,j}$ is the voxel-level predicted brain age for image $i$ and voxel $j$.  Eq. \ref{eq:global} is the global-level MAE, averaged over the batch where $y_{i}$ is the global brain age and $\hat{y}_{i}$ is the global predicted brain age for image $i$. MAE is the absolute difference between the ground truth and the predicted age. In \crefrange{eq:dice}{eq:global}, $m$ is the batch size, and $n$ is the total number of brain voxels in one sample. 

\begin{equation}
\label{eq:voxel}
\text{MAE}_{\text{voxel}}=\frac{1}{m}\sum_{i=1}^m\frac{1}{n} \sum_{j=1}^n |y^{noise}_{i,j}-\hat{y}_{i,j}| \nw{\quad \text{where} \quad y^{noise}_{i,j} = y_{i,j}+U(-2,2)}
\end{equation}

\begin{equation}
\label{eq:global}
\text{MAE}_{\text{global}}=\frac{1}{m} \sum_{i=1}^m |y_{i}-\hat{y}_{i}|
\end{equation}

The weighted sum (Eq. \ref{eq:lossfunction}) of the three terms is the loss function ($\mathcal L$) to be optimized during training. The weights $w_s$, $w_g$, and $w_v$ were set empirically and changed as the training progressed. The weight for the segmentation output, $w_s$, was initialized with the highest weight owing to the value being the smallest among the three loss terms (ranging between 0-1). The global age and voxel-wise age prediction weights, $w_g$, and $w_v$, respectively, were initialized with equal weights and updated as described in Table \ref{lossweights}. \nw{The individual weights were manually initialized and changed as the training progressed rather than being learned during training as hyperparameters to ensure the weights do not simply converge to zero to minimize the overall loss without effectively minimizing the individual loss terms. The choice of weight initializations were inspired by the range of each of the loss metrics as well as observation of the individual loss values in the initial epochs as the training progressed. This technique included experimentation to arrive at the weights proposed in the manuscript.}

\begin{equation}
\label{eq:lossfunction}
\mathcal L=  w_s  DICE_{loss} + w_v  MAE_{voxel} + w_g  MAE_{global}
\end{equation}

\begin{table}
\caption{Loss function weights as the training progresses.}
\label{lossweights}
\centering
\begin{tabularx}{0.7\linewidth}{p{60pt}XXX}
\toprule
\addlinespace[3pt]
\multirow{2}{*}{\textbf{Weight}} & \textbf{Epochs} & \textbf{Epochs} & \textbf{Epochs} \\
&$\mathbf{\in \left[ 0, 50 \right)}$ &  $\mathbf{\in \left[ 50, 130 \right)}$ &$\mathbf{\in \left[ 130, 300 \right]}$ \\
\addlinespace[3pt]
\toprule
\addlinespace[2pt]
$w_s$ & 80 & 40 & 15 \\
$w_g$ & 1 & 1 & 0.7 \\
$w_v$ & 1 & 1 & 1.3 \\
\bottomrule
\end{tabularx}
\end{table}

To ensure that the model does not learn a uniform prediction of brain age across all voxels, a subtle noise component is introduced into the loss calculation during the model's training. This \nw{uniform} noise (U), randomly selected from the range of -2 to +2, is added to the ground truth labels for each voxel (Eq. \ref{eq:voxel}). This strategic addition of noise encourages the model to learn the nuanced variations in the brain aging process across distinct regions and across subjects. We hypothesize that, when exposed to a combination of added noise and variations in underlying structural features, the model will develop accurate representations of these variations during training. By constraining the noise to a narrow range of -2 to +2, it is ensured that the effect on the training process is limited to an intentionally added randomization, without significantly impacting the model training process. \nw{The noise component influences the MAE calculation during the training loss computation and consequently, helps with a robust update of weights during back-propagation. The noise, despite being small, helps the model to learn variations in structural features as observed in the T1-weighted images by acting as a tool to guide the model towards learning features such that all voxels are not assigned the same brain age predictions, but rather, predict brain age based on small variations in structural changes as observed in the T1-weighted images.}

To evaluate the significance of incorporating noise into the ground truth labels, a variant of the proposed model without the inclusion of any additional noise was also trained. In this model, chronological age is assigned to each voxel in the ground truth labels for the voxel-level brain age prediction task based on the assumption discussed in Section \ref{sec:introduction}. In the subsequent result section, a performance comparison of both models is described.

\subsection{Ablation study}
\label{subsec:ablation}
An ablation study was performed to verify the choice of a multitask architecture. The objective is to demonstrate that both the global-brain age prediction task and the brain tissue segmentation task contribute to the model learning enhanced and accurate representations, specifically geared towards improving performance in the primary task \textit{i.e.} the voxel-level brain age prediction. Multiple models are trained, starting with a single output model that predicts voxel-level brain age, iteratively adding the other two tasks, one at a time, to analyze how models with different output tasks trained on the same dataset perform in comparison to one another. Thus, 4 different models were trained: 1) a one-task model to predict voxel-level brain age (V), 2) a two-task model to predict voxel-level brain age and segmentations of GM, WM and CSF (S+V), 3) a two-task model to predict voxel-level brain age and global-level brain age (G+V) and 4) a three-task model that predicts voxel-level brain age, global-level brain age and segmentations of GM, WM and CSF (proposed model, S+G+V).

\subsection{Network training}
\label{subsec:training}
\subsubsection{Proposed Model}
The Cam-CAN dataset was used for training the proposed model. A train:validation:test split of 489:64:98 subjects was used. Patches of size $128\times128\times128$ voxels were randomly cropped from the MR images on the fly and used as input to the model. Using patches is helpful in reducing the computational load during training allowing for the incorporation of a bigger batch size. Random cropping was done to ensure that a large majority of data samples in each batch had a significant part of the brain region, and randomizing the cropping process helps in exposing the model to brain regions from different perspectives, leading to accurate and robust features being learned.
The model was trained for 300 epochs with a batch size of 2. The Adam Optimizer was used with an initial learning rate of 0.001, weight decay of $1\mathrm{e}{-5}$, and beta values set to (0.5, 0.999). The learning rate decreased every 70 epochs by a multiplicative factor of 0.6. The hyperparameters were empirically selected.

\begin{figure}[h]
\centering
\includegraphics[width=0.9\textwidth]{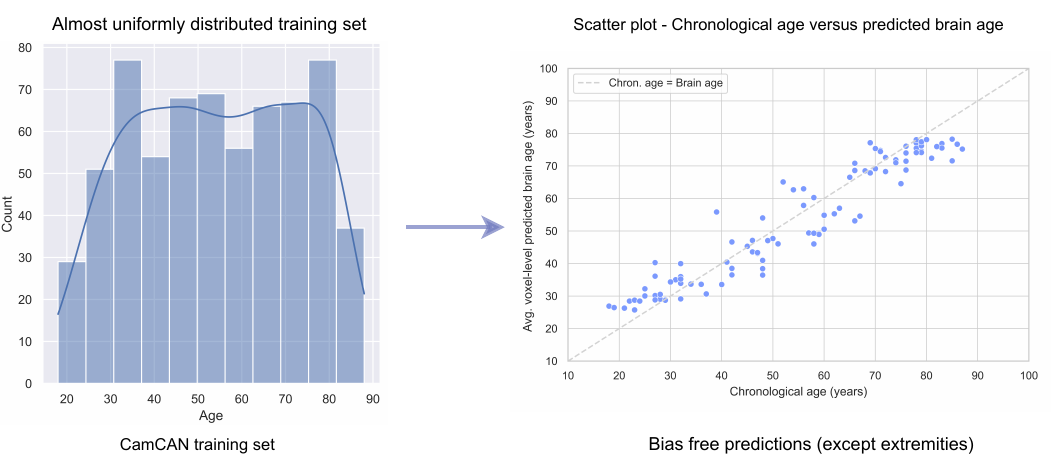}
\caption{(left) Cam-CAN training set follows a rough uniform data distribution, exposing the proposed model to samples of all ages. (right) This leads to bias-free predictions mostly, except for the extremities (ages 20-30 and ages 80-90). It can be observed that the predictions are closely aligned around the regression line for ages 30-80, with slight bias observed on the edges. A correction methodology can help correct the observed bias.}
\label{scatter_plot}
\end{figure}

\subsubsection{Ablation Study}
The same train:validation:test split of the Cam-CAN dataset used for the proposed model was used to train the ablation study models described in Section \ref{subsec:ablation}. All ablation experiment models were tested on 50 test set subjects from the Cam-CAN dataset and 359 subjects from the CC359 test set. The CC359 was split into six subsets (as described in Section \ref{subsec:data}) based on the scanner used and the magnetic field strength at which the data was acquired. Metrics were obtained for each of the six subsets to compare performance across the varying subsets. 

The one-task model to predict voxel-wise brain age and the two two-task models (segmentation/global age + voxel-wise brain age) were all trained for 300 epochs.  Various hyperparameters were experimented with. However, the most suitable ones were found to be similar to the ones used to train the proposed model, with a slight difference in the beta values that were set to default (0.9, 0.999) for the Adam optimizer. 

\subsection{Bias Correction}
\label{subsec:biascorrection}
Bias correction is a post-processing step in brain age prediction pipelines. This step is essential to remove bias due to the mean age of the training set. Brain age prediction models have been observed to be biased around the mean age of the training dataset, leading to under-estimations of brain age for subjects older than the mean age and over-estimations for subjects younger than the mean age. The source of this bias is largely unknown but is speculated to be due to reasons including noisy data, heterogeneity in the training set, data distribution, availability of data corresponding to varying age ranges, and the modeling techniques used \citep{aycheh2018biological,cole2017predicting,liang2019investigating}. A uniform dataset (Cam-CAN) during training was used, exposing the model to a balanced number of samples across all age ranges (and balanced sex distribution), minimizing biased predictions. However, despite using an \nw{ostensibly} uniform dataset, the number of samples in the extremities (20-30 years, and 80-90 years) is comparatively lower than the rest. 

The proposed methodology \nw{employed} for the bias correction technique followed what was proposed in \citet{popescu2021local}, which based on the current literature is the only study that proposed a bias correction for voxel-level brain age prediction algorithms. Hence, the goal is to train a model that learns age-specific structural features relevant to predict age such that the predictions have minimal bias. This can be confirmed by comparing the results before and after bias correction, a small difference between the two indicates that bias correction does not impact the results significantly, and hence, predictions are minimally biased.

\subsection{Regional Analysis of PAD maps}
\label{subsec:regionalanalysis}
Research in the field of brain aging studies the aging process at a regional level \textit{i.e.} in the context of different regions of the brain. To better understand the PAD maps and to assess the clinical relevance, a regional analysis of the predicted age difference at the level of known regions of the brain was performed. The publicly available MNI structural atlas \citep{collins1995automatic, mazziotta2001probabilistic} provided by the Research Imaging Center, University of Texas Health Science Center at San Antonio, Texas, USA that segments the brain into 9 anatomical regions namely Caudate, Cerebellum, Frontal Lobe, Insula, Occipital Lobe, Parietal Lobe, Putamen, Temporal Lobe and Thalamus is used. \nwf{Additionally, regional averages for the Ventricles and White Matter in the brain are also computed in the MNI space. The obtain the said averages, the MNI152 \citep{fonov2011unbiased} template was used to obtain the Ventricle and White Matter segmentation using the ``SynthSeg" tool \citep{billot2023synthseg} available as part of the FreeSurfer software package \citep{fischl2012freesurfer}.} Voxel-level brain PAD values are aggregated within each of the 11 regions to compute the average brain PAD for each region in the healthy and diseased test sets. 

\subsection{Overview from an interpretability perspective}
\label{subsec:interpretability}
Previously, with the aim of understanding regional contributions to brain age and ensuring accurate features are learned during training, global age prediction models have been explained using traditional interpretability methods. In this contribution, insights obtained from the voxel-level PAD maps are compared to the `traditional' way of understanding the models. To do so, a publicly available state-of-the-art Simple Fully Convolutional Neural network (SFCN) for global age prediction \citep{peng2021accurate,gong2021optimising} was used and three interpretability methods were implemented on it: (i) Grad-CAM \citep{selvaraju2017grad}, (ii) Occlusion Sensitivity maps \citep{zeiler2014visualizing} and (iii) SmoothGrad \citep{smilkov2017smoothgrad}. The heatmaps/saliency maps obtained were contrasted against voxel-level and regional-level PAD maps and observations were discussed. 

The SFCN model was originally designed to approach the brain age prediction task as a soft classification task, however, for the proposed implementation, the output layers of the architecture are modified to a regression head and same feature extractor is utilized as done in the original work. The Cam-CAN dataset was used to train the model following the same train:test split as done for the proposed model for fairness with the difference lying in the preprocessing of the input MR images. As the original modeling process utilized linearly registered images, the same steps were performed to linearly register the training images to the MNI template before feeding them as input to the model. An important consideration here is that no registration is performed for the proposed model, and hence the PAD maps obtained are in the native image space, whereas the interpretability heatmaps obtained are in the MNI space. Even though linear registration (or 6 degrees of freedom registration) does not alter the shape of the brain as it only implements translational and rotational changes, \nw{it can still introduce smoothening in the brain features during the process}, \nwf{which we want to avoid to retain the structural integrity of the brain. On the other hand, non-linear registration can distort the existing brain structure in aligning the MR scan to the MNI space, which is also not desired.} The uniqueness of each brain's shape and structure contributes to the prediction of brain age, and hence, it was decided against performing any registration (linear or non-linear) for the voxel-level brain age prediction model.


\begin{figure*}[h]
\centering
\includegraphics[width=\textwidth]{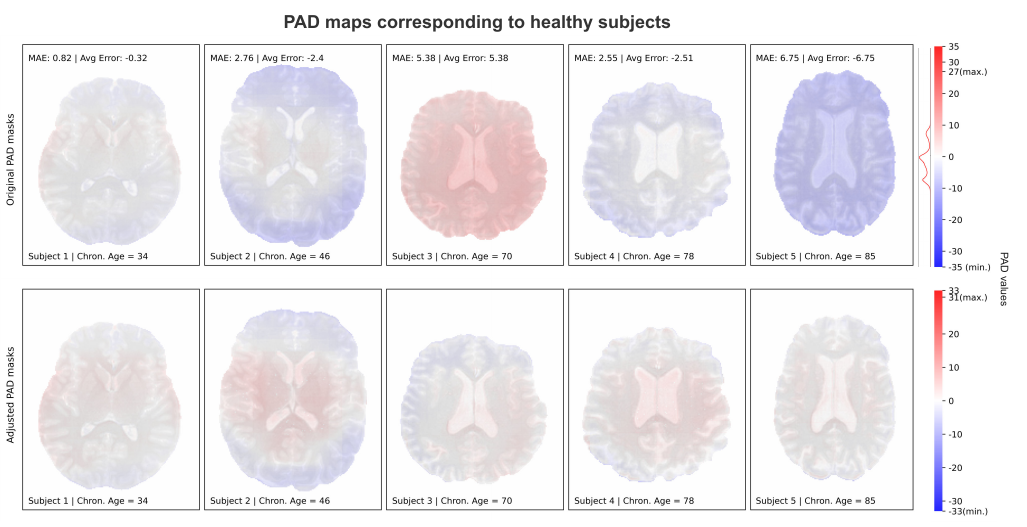}
\caption{Row 1 - PAD maps \nw{(raw, with no bias correction)} based on the voxel-level difference between chronological and predicted age, Row 2 - adjusted PAD maps by subtracting the overall MAE of the brain volume from each voxel PAD value. Each row is plotted with an independent colormap based on the range of values observed in the samples plotted in that row. \nw{Minimum and maximum points on the color bar denote the minimum and maximum voxel PAD observed across the samples in the specific row.} The data distribution plot beside the color bar in row 1 shows the distribution of PAD values across the entire healthy test set. Extended analysis of healthy PAD maps is described in \citep{gianchandani2023}.}
\label{pads_Cam-CAN}
\end{figure*}

\section{Results}
\label{sec:results}
For a fair comparison of model performance and as suggested in \citet{popescu2021local}, all results are reported before bias correction. Bias-corrected results are only used for visualizations and analysis of diseased subjects where explicitly stated. \nw{In Table \ref{tab:results_main}, columns 2 and 3 with header voxel-level MAE refer to averaged voxel-level prediction results, and columns 4 and 5 with header global MAE refer to the global brain age prediction results. Voxel-level MAE results are considered for model comparisons throughout the work.}

\begin{figure*}[h]
\centering
\includegraphics[width=\textwidth]{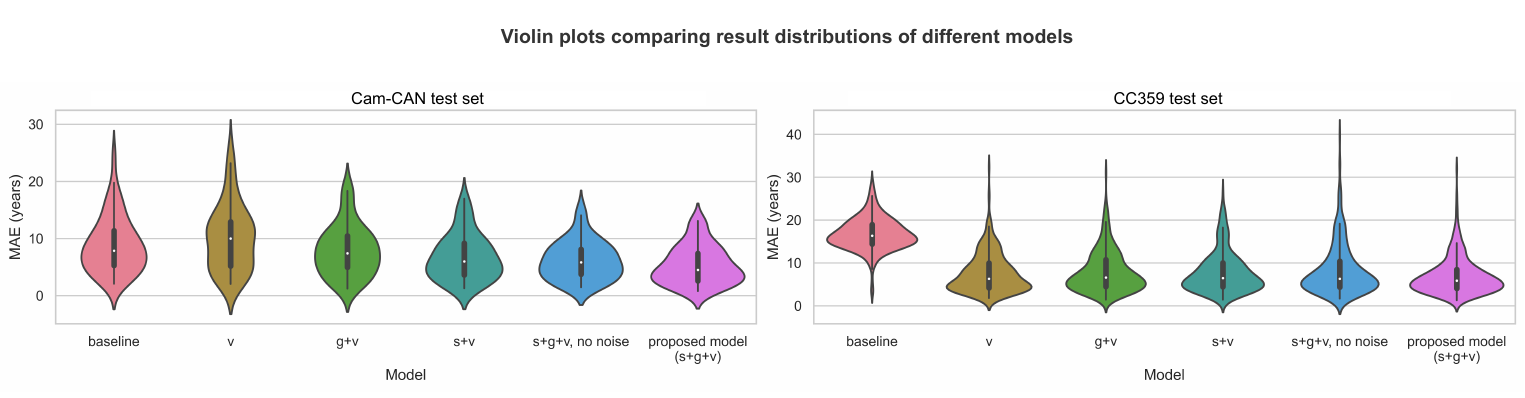}
\caption{\nw{Violin plots comparing the distribution of test results (voxel-level MAE) of different models. Each violin represents the probability density of data, with quartiles displayed as black lines and the median indicated by the white dot. It can be observed that for both the Cam-CAN and CC359 test sets, the median MAE is smaller for the proposed model compared to the baseline. All results are reported before bias correction.} }
\label{violin_plots}
\end{figure*}

\begin{figure*}[h]
\centering
\includegraphics[width=\textwidth]{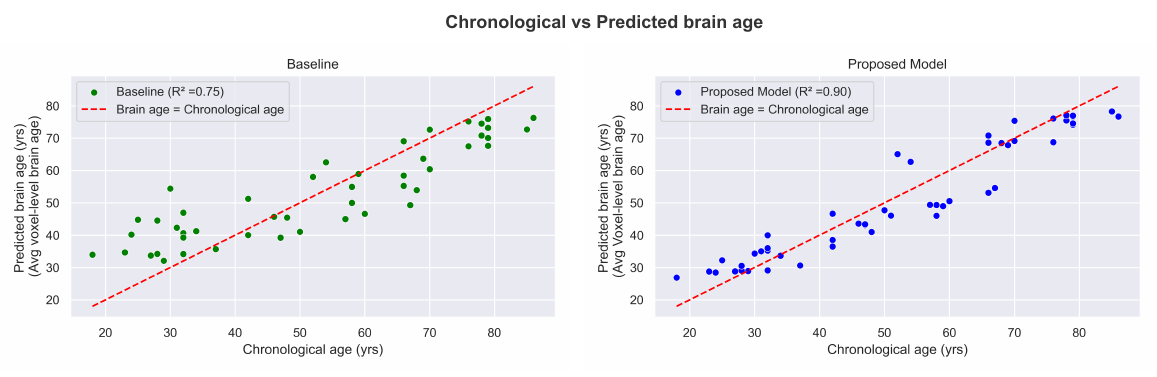}
\caption{\nw{The scatter plots show the predicted versus chronological age for the baseline and proposed model. The baseline achieved an $R^2$ of 0.75 whereas the proposed voxel-level brain age prediction model achieved an $R^2$ of 0.90 on the Cam-CAN test set.} }
\label{r_square}
\end{figure*}

\textbf{Contribution 1: Proposal of a multitask DL voxel-level brain age prediction model: }
The proposed model surpasses the baseline (refer Table \ref{tab:results_main} - Voxel-level MAE), demonstrating a 39.22\% reduction in MAE on the internal Cam-CAN test set. The proposed model is also evaluated on a larger external test set (CC359) and obtains an MAE reduction of 58.88\% which reflects on the model's performance on unseen data originating from a different data source. The proposed model variant (with 3-output) without added noise to the loss function comes in second on the Cam-CAN evaluation and second to last on the CC359 test set. \nw{The global level MAE is reported for the models that included global brain age predictions in Table \ref{tab:results_main} - Global MAE column. It can be observed that the SFCN model outperformed the remaining models on the Cam-CAN test set, whereas the proposed model achieved the smallest MAE on the CC359 test set, however, it must be noted that this work aims to predict voxel-level brain age and global brain age was added as a complementary task in the proposed model for improved feature extraction for the main voxel-level brain age prediction task. Violin plots showing the data distribution of voxel-level test results can be observed in Figure. \ref{violin_plots}. The median MAE is seen to be lower for the proposed model as compared to the baseline on both test sets. It can also be observed that variation in model performance is almost comparable for all models, however, for both test sets, the widest part of the violin that indicates higher variability or spread in MAE is seen at a lower error value for the proposed model as compared to the baseline. Figure \ref{r_square} shows the scatter plots for the baseline and proposed model's voxel-level brain age prediction results with respective $R^2$ scores (coefficient of determination) that show the goodness of fit of a model.}

\begin{figure*}[h!]
\centering
\includegraphics[width=\textwidth]{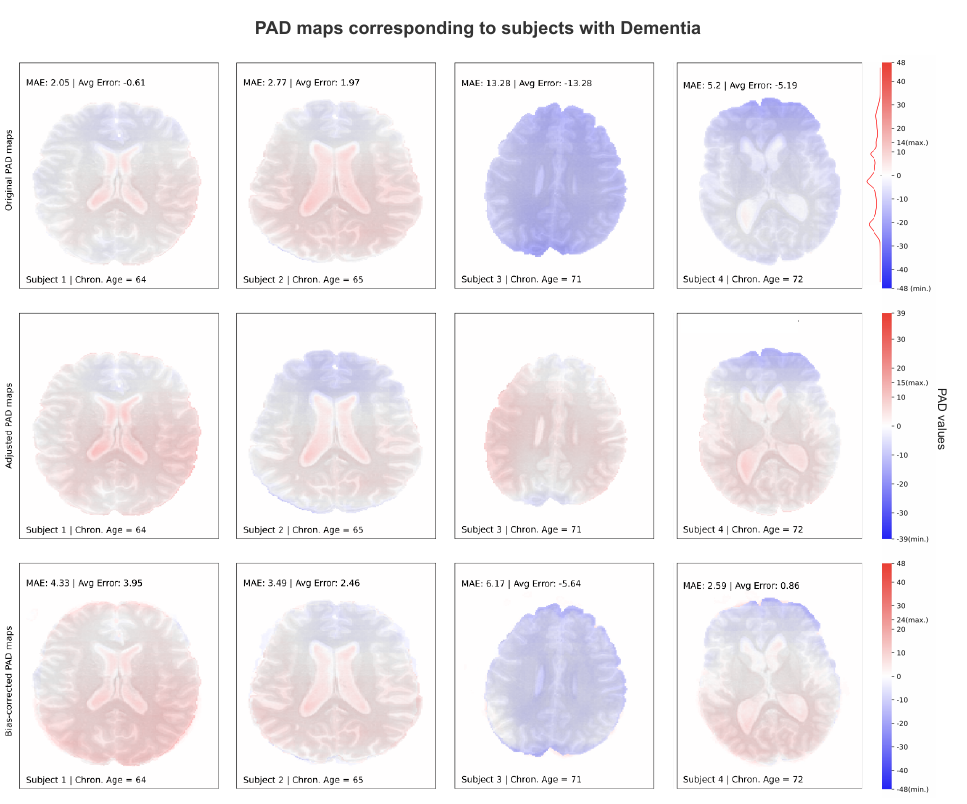}
\caption{PAD maps corresponding to diseased subjects (OASIS dataset). Row 1 shows the raw PAD maps obtained from the voxel-level brain age prediction model. Row 2 shows the adjusted PAD maps (for improved visualization). Row 3 shows bias-corrected PAD maps using the correction methodology described in Section \ref{subsec:biascorrection}. More red regions are observed as compared to healthy PAD maps and accelerated aging in the ventricles which has often been associated with neurological disorders. Each row is plotted with an independent colormap based on the range of values observed in the samples plotted in that row. \nw{Minimum and maximum points on the color bar denote the minimum and maximum voxel PAD observed across the samples in the specific row.} The data distribution plot beside the color bar in row 1 shows the distribution of PAD values across the entire test set of subjects with dementia. A wider spread of PAD values is observed compared to the data distribution of the healthy test set.}
\label{oasis_diseased}
\end{figure*}

\begin{table}[h!]
    \centering
    \begin{threeparttable}
        \caption{Model performance on an internal (Cam-CAN) and external test set (CC359). \nw{All results are reported before bias correction.} }
        \label{tab:results_main}
        \begin{tabular}
{p{0.38\textwidth}p{0.12\textwidth}p{0.12\textwidth}p{0.12\textwidth}p{0.11\textwidth}}
            \toprule
            \textbf{Model (output tasks)} & \multicolumn{2}{c}{\textbf{Voxel-level MAE}} & \multicolumn{2}{c}{\textbf{Global MAE}} \\
            \cmidrule(r){2-3} \cmidrule(){4-5}
            &  Cam-CAN &  CC359 &  Cam-CAN &  CC359 \\
            \midrule
            Global age (G) & - & - & 5.32$\pm$3.67 & 6.50$\pm$4.71\\
            Baseline (G+V) & 8.84$\pm$4.82 & 16.74$\pm$3.71 & - & - \\
            1 output model (V) & 10.11$\pm$5.68 & 7.63$\pm$4.53 & - & - \\
            2 output model (G+V) & 7.90$\pm$4.30 & 7.93$\pm$4.73 & 6.61$\pm$4.17& 6.52$\pm$5.22 \\
            2 output model (S+V) & 6.75$\pm$3.94 & 7.83$\pm$4.74 & - & - \\
            3 output model (S+G+V), no noise & 6.14$\pm$3.32 & 8.32$\pm$5.84 & 5.83$\pm$3.98& 8.70$\pm$7.16\\
            \textbf{Proposed model (S+G+V)} & \textbf{5.30$\pm$3.29}\textsuperscript{*} & \textbf{6.92$\pm$4.28}\textsuperscript{*} & 6.11$\pm$3.80 & 5.51$\pm$4.38\\
            \bottomrule
        \end{tabular}
        \smallskip
        \begin{tablenotes}
            \small  
            \item \textbf{Abbreviations}: V - voxel-level brain age prediction task, S - segmentation task (GM, WM, CSF), G - global-level brain age prediction task, \textsuperscript{*}- p\textless{}0.05
        \end{tablenotes}
    \end{threeparttable}
\end{table}

For voxel-level predictions, since it is impossible to present prediction results at the level of each voxel (millions in each brain volume), the mean of the per-sample MAE ($\text{MAE}_{\text{voxel}}$) is reported in Table \ref{tab:results_main}. To visualize the voxel-level brain age predictions, predicted age difference (PAD) maps are used, which show the difference between the predicted brain age and the chronological age at the level of each voxel. PAD maps for the Cam-CAN test set samples can be observed in Figure \ref{pads_Cam-CAN}, where blue color indicates brain regions that look younger than chronological age and red correlates to older-looking brain regions. The first row corresponds to the raw PAD maps whereas the second row corresponds to the adjusted PAD maps obtained by subtracting the overall MAE of the brain volume from each voxel PAD value. These adjusted maps allow us to visualize the spatial variations in PAD values across different regions of the brain without the interference of the model error (MAE). The adjusted PAD maps are constructed purely for visualization purposes and are not used for any result comparisons with other models/baseline. Similarly, the PAD maps corresponding to subjects with dementia can be observed in Figure \ref{oasis_diseased}. At a high level, it can be observed from the PAD maps corresponding to healthy versus dementia subjects, that the contrasts are sharper and more apparent in subjects with dementia reflecting greater variation in regional brain ages. Additionally, the PAD maps for subjects with dementia have intensity PADs spread across a wider range of values, which can be observed from the distribution of values shown alongside the color bar \nw{in row 1 } in Figure \ref{oasis_diseased} as well more red regions as compared to healthy PAD maps. More analysis on healthy PAD maps is done in \citet{gianchandani2023} and that on diseased subjects will be further discussed in the subsequent sections.

The Wilcoxon-Signed Rank test was performed to assess the voxel-level performance of the proposed model against other variations (1-output, 2-output) of the model and the baseline. \nw{Hence, 5 statistical tests were performed, the proposed voxel-level brain age prediction model against each of the models in rows 2-6 of Table \ref{tab:results_main}.} $\alpha$  was set to 0.05 and the Holm-Bonferroni correction was done to account for multiple comparisons. All resulting p-values were found to be less than $0.05$, indicating statistical significance.

\nw{The proposed model was also tested on the CC359 dataset stratified by sex. The model performed with a difference of $\sim$ 1.2 year in the voxel-level MAEs on the two test sets. The proposed model achieved an MAE of $7.54\pm4.78$ years on the $\text{CC359}_{male}$ test set (n=176) and an MAE of $6.32\pm3.61$ years on the $\text{CC359}_{female}$ test set (n=183). }

\begin{table}[h!]
\centering
\begin{threeparttable}
\caption{Ablation study results. All results are reported before bias correction.}
\label{ablation}
\begin{tabularx}{0.9\linewidth}{p{118pt}p{135pt}X}
\toprule
\textbf{Test Set} & \textbf{Model (output)} & \textbf{MAE$\pm$S.D.} \\
\midrule
\multirow{4}{*}{Philips 1.5T} & V & 7.22$\pm$3.13 \\
& S+V & 7.83$\pm$4.63 \\
& G+V & 9.20$\pm$5.36 \\
& S+G+V (proposed) & \textbf{6.94$\pm$3.80}\textsuperscript{* S+V,G+V} \\
\midrule
\multirow{4}{*}{Philips 3T} & V & 8.02$\pm$5.29 \\
& S+V & 9.61$\pm$5.33 \\
& G+V & 9.54$\pm$5.99 \\
& S+G+V (proposed) & \textbf{7.73$\pm$5.04}\textsuperscript{* S+V,G+V} \\
\midrule
\multirow{4}{*}{Siemens 1.5T} & V & 8.26$\pm$5.33 \\
& S+V & 8.64$\pm$5.60 \\
& G+V & 8.75$\pm$4.37 \\
& S+G+V (proposed) & \textbf{6.68$\pm$4.80}\textsuperscript{* V, S+V,G+V} \\
\midrule
\multirow{4}{*}{Siemens 3T} & V & 9.18$\pm$5.29 \\
& S+V & 6.21$\pm$4.13 \\
& G+V & \textbf{5.84$\pm$4.10}\textsuperscript{* V, S+G+V} \\
& S+G+V (proposed) & 6.80$\pm$4.22 \\
\midrule
\multirow{4}{*}{GE 1.5T} & V & 5.83$\pm$3.21 \\
& S+V & 7.17$\pm$3.51 \\
& G+V & \textbf{5.79$\pm$2.34}\textsuperscript{* S+V} \\
& S+G+V (proposed) & 5.98$\pm$2.52 \\
\midrule
\multirow{4}{*}{GE 3T} & V & \textbf{7.26$\pm$3.46}\textsuperscript{* G+V} \\
& S+V & 7.55$\pm$4.08 \\
& G+V & 8.49$\pm$3.73 \\
& S+G+V (proposed) & 7.40$\pm$4.52 \\
\bottomrule
\end{tabularx}
\smallskip
\begin{tablenotes}
\small
\item 1. All test sets have n=60 samples, except Philips 1.5T with n=59 samples. 
\item 2. Abbreviations: V - voxel-level brain age prediction task, S - segmentation task (GM, WM, CSF), G - global-level brain age prediction task.
\item 3. Statistical significance is shown with the symbols, V, S+V, G+V or S+G+V which denote each model. The symbol beside an MAE value indicates statistical significance (with p\textless{}0.05) with the specific model mentioned.
\end{tablenotes}
\end{threeparttable}
\end{table}

\textbf{Contribution 2: Ablation study to show the importance of using a multitask architecture: }
As stated in Section \ref{subsec:ablation}, the proposed three-task (multitask) model is expected to show superior performance on the voxel-level brain age prediction task compared to the one-task and two-task counterparts. An ablation study is performed by designing experiments with the same model architecture with different task combinations, and it can be observed in Table \ref{tab:results_main}, that the 3-output proposed model outperforms the 1-output and 2-output models with statistically significant results (p\textless{}0.05) on the internal Cam-CAN test set. To further validate the findings, all ablation study models are subjected to evaluation using the CC359 dataset. This dataset comprises data acquired from 3 distinct scanner vendors, each acquired at 2 different magnetic field strengths. Consequently, this dataset is segregated into 6 subsets, all sharing similar acquisition protocols. The evaluation is conducted independently on each subset (refer to Table \ref{ablation}) for every ablation experiment model. It is observed that the proposed model outperforms the 1-output \nw{(V)} and 2-output models \nw{(S+V, G+V)} on 3 out of 6 subsets (Philips 1.5T, Philips 3T, Siemens 1.5T), comes close second on 1 subset (GE 3T) and takes the third spot on the 2 subsets (GE 1.5T, and Siemens 3T). Closely inspecting the subsets where the proposed model did not take the lead, it was observed that for the GE 3T subset, the proposed mode ranked second with an average MAE on the test set differing by no more than 0.2 years. Similarly, on GE 1.5T and Siemens 3T subsets, where the proposed model secured the third position, the difference between the top-ranking model and the proposed three-task model was at best 1 year.

\nw{The Wilcoxon-Signed Rank test was performed to assess the statistical significance of the results. For each subset of CC359, the winning model was compared to the remaining three models, and multiple comparisons were accounted for by performing the Holm-Bonferroni correction. For each test set, if the winning model was found to have significant results against another model, the symbol assigned to the model has been shown beside the MAE value in column 3. It can be observed that even though the proposed model obtained significantly better results against all the other model variants only on the Siemens 1.5T test set, the performance is consistently better than any other variants of the proposed model. The 1-output \nw{(V)} and 2-output \nw{(S+V, G+V)} models only performed significantly better than 1 or 2 other models whereas the proposed model consistently had significantly better results than 2 or 3 of the other model variants.}

Overall, the proposed model outperformed the ablation experiment models on 50\% of the subsets, while consistently performing well across all subsets, unlike the 1-output and 2-output models which obtained significantly higher errors ($\sim$9 years) on at least 1 or more of the subsets. The proposed model consistently achieved an average MAE in the range of 5.9 to 7.7 years across all subsets of CC359, whereas other ablation experiment models (1-output and 2-output) exhibited greater fluctuations in the inter-dataset performance. Evaluation on subsets acquired using different scanners, which in turn exhibit scanner-specific differences in the MR images, and at different magnetic field strengths reflects on the model's ability to be robust and generalizable across diverse datasets. 

\begin{figure}[h]
\centering
    \subfigure[]
    {
    \includegraphics[width=\textwidth]{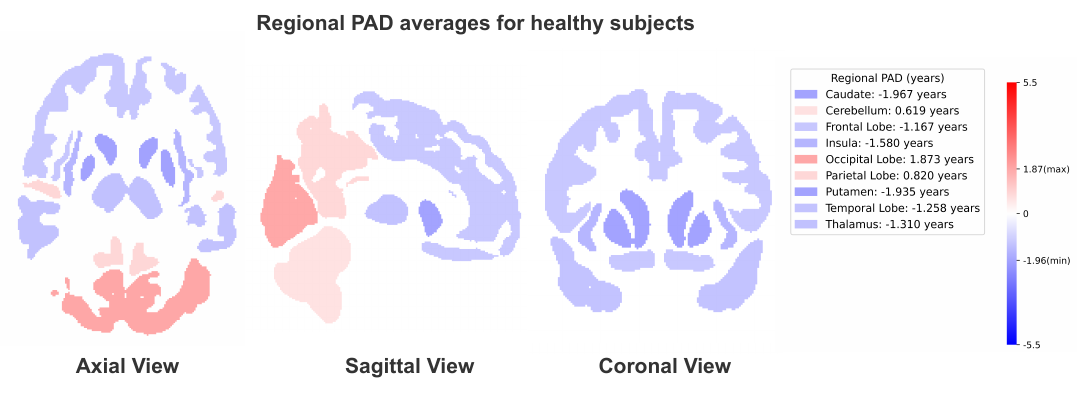}
    }
    \subfigure[]
    {
        \includegraphics[width=\textwidth]{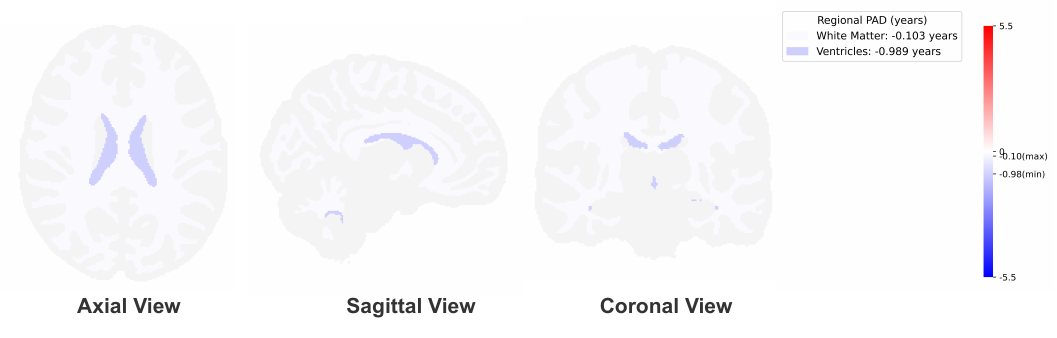}
    }
\caption{\nw{Regional PAD averages for different regions of the brain in a population of presumed healthy subjects. The visualization is in MNI space and has been created using 40 unseen subjects during training from the Cam-CAN dataset. Bias correction as well as filtering for subjects$\leq$70 years of age is done to match the AD and dementia test sets used for regional analysis. It can be observed that most regions of the brain show small negative PAD values with the Caudate looking the youngest with a -1.96 years PAD. Three regions showed positive PAD with Occipital Lobe exhibiting the largest PAD at 1.87 years. \nwf{Average regional age of the ventricles was -0.98 years making them appear younger than usual.}}}
\label{mni_atlas_Cam-CAN}
\end{figure}

\begin{figure}[h!]
\centering
 \subfigure[]
    {
    \includegraphics[width=\textwidth]{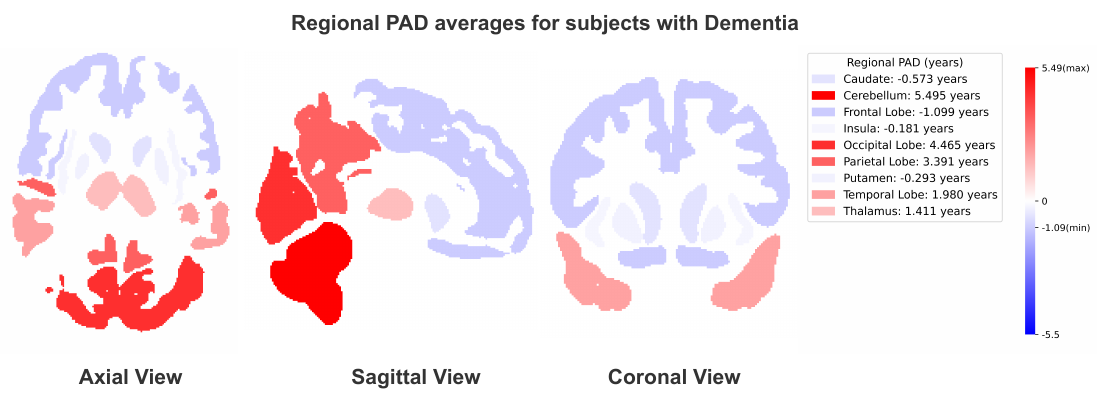}
    }
     \subfigure[]
    {
    \includegraphics[width=\textwidth]{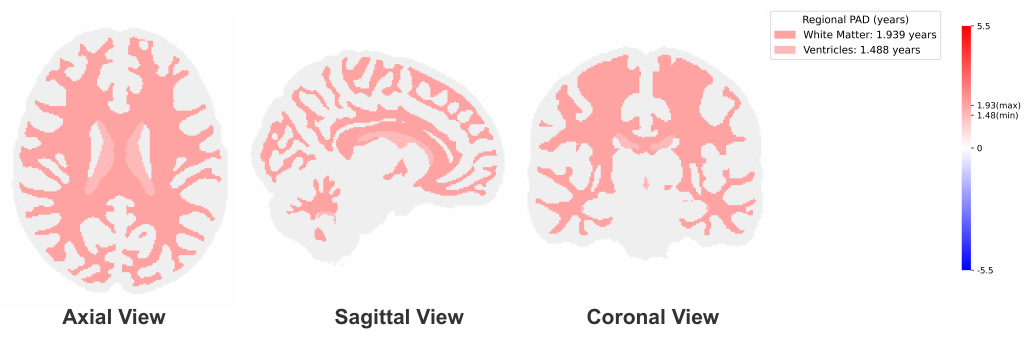}
    }
\caption{\nw{Regional PAD averages for different regions of the brain in a population of subjects with dementia from the OASIS dataset. \nw{The visualization is constructed with bias-corrected PAD maps.} The visualization is in MNI space }and has been created using 15 subjects with age $\leq$ 70 years using the voxel-level bias-corrected PAD maps. A variation is observed in terms of PAD values across different regions with the Cerebellum, Occipital Lobe, Parietal Lobe, Temporal Lobe, and Thalamus showing an increased brain age \nw{(largest PAD of 5.4 years for Cerebellum. \nwf{Ventricles show an increased brain age with an average regional PAD of 1.48 years.})}}
\label{mni_atlas_Dementia}
\end{figure}

\begin{figure}[h!]
\centering
    \subfigure[]
    {
    \includegraphics[width=\textwidth]{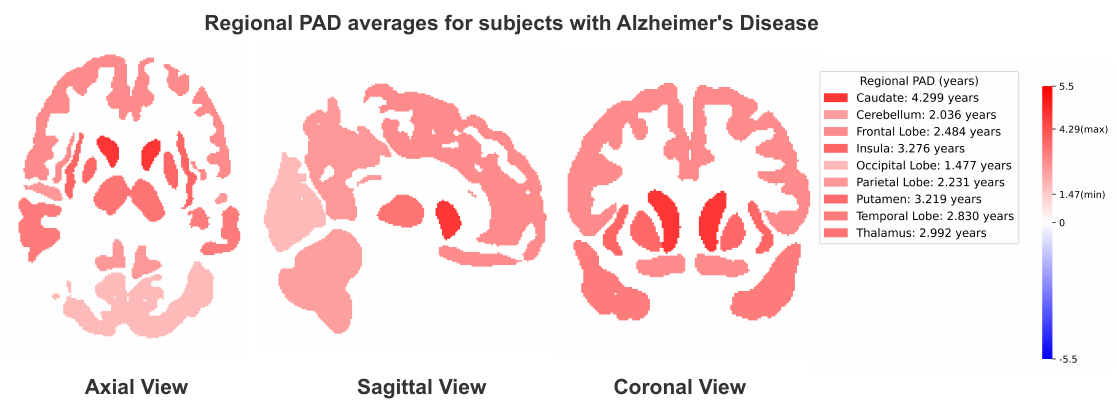}
    }
    \subfigure[]
    {
    \includegraphics[width=\textwidth]{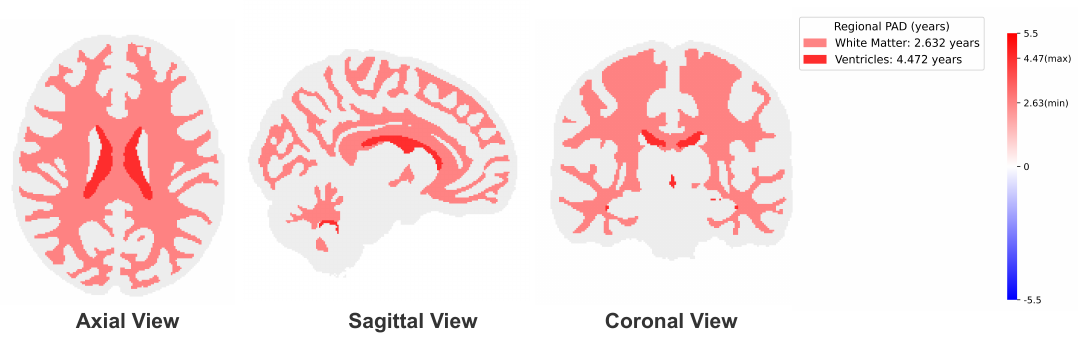}
    }
    
\caption{\nw{Regional PAD averages for different regions of the brain in a population of subjects with AD from the ADNI dataset. \nw{The visualization is constructed with bias-corrected PAD maps.} The visualization is in MNI space and} has been created using 17 subjects with age $\leq$ 70 years using the voxel-level bias-corrected PAD maps. It can observed that all regions in the atlas show an increased brain age \nwf{with the largest PAD observed in the Ventricles at 4.47 years closely followed by Caudate at 4.29 years.}}
\label{mni_atlas_alzheimers}
\end{figure}

\textbf{Contribution 3: Regional analysis of the brain aging process in a healthy versus diseased brain: }
The proposed model was tested on healthy subjects from the Cam-CAN dataset, which was used for the regional analysis. For the evaluation of diseased subjects, subjects with AD from the ADNI dataset (n=20) and subjects with dementia from the OASIS3 dataset (n=28) were utilized. It is essential to note that the majority of the open-source MR images of subjects with neurological disorders (especially AD and dementia) correspond to older age ranges, usually 55 years and above with the frequency of samples available increasing as one goes higher up. To mitigate any biased predictions, filtering was performed on all test sets (healthy, AD, and, dementia) for subjects with age $\leq$ 70 years for the regional analysis, leaving us with \nw{n=40 healthy subjects} and n=32 subjects \nw{with either AD or dementia for the analysis.} This decision will be further justified in the discussion section.

In Table \ref{tab:regionalpad}, the regional PAD average and standard deviation (S.D.) values based on the MNI structural atlas (refer to section \ref{subsec:regionalanalysis}) are reported. The regional analysis on three test sets, one corresponding to healthy subjects (Cam-CAN) and two diseased test sets (AD and dementia) was performed. For each dataset, the average (Mean $\pm$ S.D.) PAD values for each region across the test set samples were reported. Additionally, S.D. per region is described (Mean of S.D. $\pm$ S.D. of S.D.) to observe the variability of PAD values within independent regions. \nw{To check for statistically significant differences between test sets for each regional average PAD, the Kruskal-Wallis test for performed. This non-parametric test was chosen as the distribution of values was not found to be normal across all populations. Following the Kruskal-Wallis test, Dunn's test was performed for pairwise comparisons between the possible test set pairs, and multiple comparisons were accounted for using the Holm-Bonferroni correction. For each region, if a significant difference was observed with p\textless{}0.05, a \textsuperscript{*} is shown beside the average regional PAD for the test set pair for each region (Table \ref{tab:regionalpad}).}

Figure \ref{mni_atlas_Cam-CAN} and \Cref{mni_atlas_Dementia,mni_atlas_alzheimers} show \nw{regional averages of PAD values} on the healthy and diseased test sets respectively. A clear distinction can be observed between the healthy versus diseased \nw{population averages} with the healthy \nw{averages} appearing to have regional PAD values closer to 0, indicating only a small deviation \nw{(less than 2 years)} from the chronological age of the subjects. In the \nw{averages} for diseased subjects (\Cref{mni_atlas_Dementia,mni_atlas_alzheimers}), red colors are observed in most regions of the brain. Overall, \nw{the regional averages for subjects with either dementia or AD} display an accelerated aging trajectory as well as sharper contrasts as compared to the regional averages corresponding to healthy subjects. \nwf{It must also be noted that the ventricular region, which is often enlarged as a result of abnormal aging in humans, appears to show an accelerated aging trajectory for both the dementia and AD populations.} \nw{To account for the differences in the orientation of different samples in a population, the PAD maps were registered to the MNI space to compute the regional average visualizations as shown in \Cref{mni_atlas_Cam-CAN,mni_atlas_Dementia,mni_atlas_alzheimers}.}

\begin{table*}[h]
\small
\caption{Regional PAD values. \nw{The analysis is done using bias-corrected voxel-level PAD maps for all test sets for consistency.}}
\label{tab:regionalpad}
\centering
\begin{threeparttable}
\setlength{\tabcolsep}{2.3pt}
\begin{tabularx}{\linewidth}{p{45pt}p{65pt}*{3}{X}p{67pt}X}
\toprule
\addlinespace[3pt]
\multirow{3}{*}{\textbf{Regions}} & \multicolumn{6}{c}{\textbf{Test sets}} \\
\cline{2-7}
\addlinespace[3pt]
 & \multicolumn{2}{c}{\textbf{Healthy}} & \multicolumn{2}{c}{\textbf{AD}} & \multicolumn{2}{c}{\textbf{Dementia}} \\
 & \centering Avg regional \\PAD & Regional S.D. & \centering Avg regional PAD & Regional S.D. & \centering Avg regional\\ PAD & Regional S.D. \\
\addlinespace[3pt]
\toprule
\addlinespace[2pt]
Caudate& \nw{$-1.66\pm7.85\textsuperscript{*}$} & \nw{$1.27\pm0.63$} & \nw{$4.28\pm4.49\textsuperscript{*}$} & \nw{$1.65\pm0.56$} & \nw{$-0.58\pm11.23$} & \nw{$1.39\pm0.45$} \\
Cerebell-um& \nw{$0.38\pm9.65$} & \nw{$3.61\pm1.64$} & \nw{$2.10\pm5.54$} & \nw{$3.60\pm1.37$} & \nw{$\quad5.44\pm10.80$} & \nw{$3.42\pm1.04$}\\
Frontal Lobe& \nw{$-1.15\pm7.04\textsuperscript{*}$} & \nw{$2.78\pm0.91$} & \nw{$2.40\pm3.92\textsuperscript{*}$} & \nw{$3.12\pm0.69$} & \nw{$-1.14\pm9.55$} & \nw{$3.61\pm1.47$}\\
Insula&  \nw{$-1.56\pm8.05\textsuperscript{*}$} & \nw{$1.70\pm0.94$} & \nw{$3.29\pm3.97\textsuperscript{*}$} & \nw{$1.54\pm0.51$} & \nw{$-0.21\pm11.14$} & \nw{$1.95\pm0.87$}\\
Occipital Lobe& \nw{$1.46\pm8.03$} & \nw{$2.89\pm1.70$} & \nw{$1.49\pm6.13$} & \nw{$2.53\pm0.87$} & \nw{$\quad 4.37\pm11.83$} &  \nw{$2.40\pm0.90$}\\
Parietal Lobe& \nw{$0.54\pm7.56$} & \nw{$2.61\pm1.07$} & \nw{$2.16\pm5.15$} & \nw{$3.11\pm0.72$} & \nw{$\quad 3.31\pm10.69$} & \nw{$3.19\pm1.36$} \\
Putamen& \nw{$-1.90\pm8.17\textsuperscript{*}$} & \nw{$1.22\pm0.57$} & \nw{$3.26\pm3.91\textsuperscript{*}$} & \nw{$1.16\pm0.38$} & \nw{$-0.31\pm11.20$} & \nw{$1.19\pm0.44$} \\
Temporal Lobe& \nw{$-1.09\pm7.52\textsuperscript{*}$} & \nw{$3.87\pm1.80$} & \nw{$2.82\pm3.12\textsuperscript{*}$} & \nw{$3.07\pm1.00$} & \nw{$\quad 1.96\pm9.90$} & \nw{$3.72\pm1.43$}\\
Thalamus& \nw{$-1.28\pm8.46\textsuperscript{*}$} & \nw{$1.05\pm0.56$} & \nw{$2.99\pm3.69\textsuperscript{*}$} & \nw{$1.06\pm0.55$} & \nw{$\quad 1.41\pm11.10$} & \nw{$1.01\pm0.32$}\\
\bottomrule
\end{tabularx}
\begin{tablenotes}
\small  
\item \textbf{Note:} Statistically significant difference (with p\textless{}0.05) between test sets for each region is shown with a \textsuperscript{*}. For example, for Caudate, there is a statistically significant difference between avg. regional PAD for Healthy and AD test sets.
\end{tablenotes}
\end{threeparttable}
\end{table*}

\textbf{Contribution 4: Interpretability analysis and comparison with traditional interpretability methods: }
PAD maps obtained from the voxel-level brain age prediction model are compared to the heatmaps obtained from 3 interpretability methods. It is imperative to note that for the scope of this article, the objective of this research is not to propose a state-of-the-art global age prediction model to obtain interpretability maps using traditional methods, however, the aim is to observe the difference in underlying properties and insights obtained from PAD maps versus traditional interpretability heatmaps.

\begin{figure*}[h!]
\centering
\includegraphics[width=\textwidth]{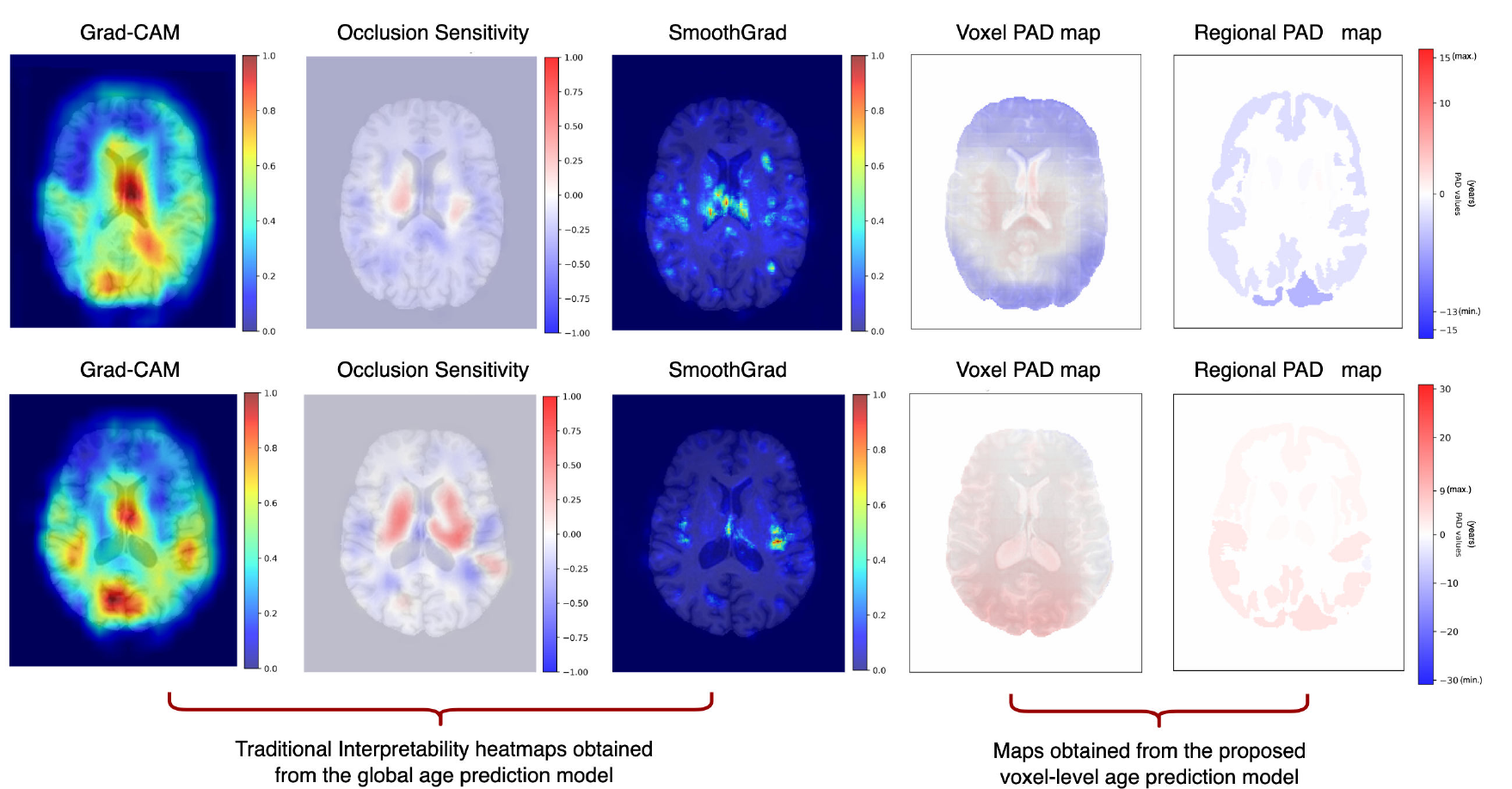}
\caption{Comparison of traditional interpretability heatmaps (left to right: Grad-CAM, Occlusion Sensitivity, and SmoothGrad) with PAD maps (left to right: voxel-level and regional) obtained from the proposed voxel-level brain age prediction model \nw{on presumed healthy subjects.} \nw{The PAD maps (columns 4 and 5) are non bias-corrected. Grad-CAM and SmoothGrad heatmaps are normalized between 0 to 1. Occlusion Sensitivity maps are bidirectional and hence, normalized between -1 to +1. PAD maps are plotted based on actual PAD values in years.}}
\label{fig:interpretability}
\end{figure*}

In Figure \ref{fig:interpretability}, the first column shows Grad-CAM heatmaps that illustrate regions with relative contribution/importance to the brain age prediction. It is often visualized using red-yellow-blue heatmaps with red regions as the most important and blue being the least. However, since Grad-CAM heatmaps are obtained from the later convolutional layers in a model to observe the final features learned through the gradient with respect to input, they are originally obtained at a much smaller size as compared to input and have to be upsampled, which leads to interpolation errors and coarse maps. The second column shows occlusion sensitivity maps where red regions make the model overestimate the brain age prediction and blue ones make the model underestimate the predictions. White regions contribute the least. SmoothGrad maps are similar to Grad-CAM heatmaps, except they are generated as a result of multiple forward passes of noisy input through the model to obtain heatmaps that are more precise, counteracting the influence of noise. However, similar to Grad-CAM they are based on the gradients with respect to an input and hence, illustrate the relative importance of regions in one input and are not comparable across samples. \nw{Traditional interpretability heatmaps (Grad-CAM, Occlusion Sensitivity maps, and SmoothGrad) maps were originally obtained in the image space registered with 6 degrees of freedom to MNI space, however, for better visualization, the images have been reverted back into the original image space to match the PAD maps in Figure \ref{fig:interpretability}.}

Voxel-level PAD maps show the regions with an increased brain age in red and decreased brain age in blue. The maps were obtained at the same resolution as the input image due to the upsampling in the U-Net architecture. The use of skip connections in the U-Net architecture leads to accurate upsampling at a high resolution.  The intensity values in the PAD maps are quantified in years by computing the difference between predicted and chronological age, and hence, are comparable across samples. The last column in the figure shows the regional PAD maps (PAD values averaged within different known anatomical regions of the brain), which essentially have similar features and characteristics as the voxel-level PAD maps with the difference being in the granularity of the PAD values. This representation, however, is better suited to analyze the results from the voxel-level age prediction model from an aging perspective.


\section{Discussion}
\label{sec:discussion}
The proposed voxel-level brain age prediction model outperforms the baseline \nw{U-net voxel-wise prediction model} on two independent test sets \nw{achieving an error reduction of greater than 30\% on both} while having a simple and straightforward preprocessing pipeline. Diverging from the baseline \citep{popescu2021local}, the proposed methodology, \nw{initially introduced in our previous work \cite{gianchandani2023}}, presents two significant modifications. First, the baseline uses non-linear registration as a pre-processing step, registering all T1-weighted images to the MNI \nw{template}, an average atlas representative of a healthy brain. We hypothesized that each brain structure is unique in terms of shape, size, and structural features and the uniqueness is crucial for brain age estimation. Non-linear registration can modify the uniqueness that each brain volume holds and information is lost in the process. \nw{Linear registration can also introduce small smoothening effects to the features of the MR image during translational and rotational changes.} Following the same, non-registered images are used as input to the proposed model. This helps retain the original shape, size, and structural features in the truest form possible to be used to predict voxel-level brain age. Second, the baseline uses GM and WM masks obtained from the non-linearly registered images as input to the model, \textit{i.e.}, whole T1-weighted volumes are not fed into the network. \nw{The authors rely on implicit CSF features embedded into WM and GM boundaries without actually incorporating CSF segmentation masks. Previous research has shown the relevance of CSF in studying the brain aging process \citep{houston2023aging,may1990cerebrospinal}. \citet{yamada2023aging} have shown the increase in intracranial CSF volume as a result of brain volume reduction with age in healthy subjects as well increase in ventricular CSF volume after 60 years of age. Composition changes in CSF volume have also been shown to be associated with AD as well as healthy aging \citep{fjell2010brain}. This indicates that CSF volume and composition changes are important indicators to consider when predicting brain age. Hence, instead of relying on the implicitly embedded CSF features in GM and WM masks,} the proposed methodology utilizes skull-stripped T1-weighted volumes \nw{that include} GM, WM, and CSF as input to the model. Segmentation of GM, WM, and CSF is added as one of the output tasks to the proposed model which also contributes to the interpretability analysis. \nw{The experiments in this work are designed to present an extended evaluation of our recently proposed voxel-level brain age prediction model using a multi-task approach \citep{gianchandani2023}. This was done through an evaluation on subjects with underlying neurological disorders, a regional analysis of voxel-level brain age predictions, and an interpretability analysis.}

To ensure the learning of accurate feature representations, a subtle noise component \nw{(uniform noise between -2 and +2)} is introduced to the ground truth labels during model training (refer to Section \ref{subsec:loss}). This strategic addition of noise serves to facilitate the model's ability to discern and understand variations in aging patterns across different brain regions. While this approach introduces noise at the voxel level, it is important to acknowledge that in certain instances, this technique could theoretically yield drastic differences in PAD values between adjacent voxels. For instance, the inclusion of noise might lead to stark contrasts, such as a red voxel (increased brain age) right adjacent to a contrasting blue voxel (reduced brain age) making the PAD mask appear with a salt and pepper noise appearance.  Despite the possibility of sharp contrasts, the PAD maps consistently reveal a tendency toward producing smooth transitions in the brain PAD values with clusters of voxel exhibiting similar patterns of aging. This phenomenon aligns with the inherent nature of aging-related changes, which tend to present on a regional level. Even though the proposed model with intentionally introduced noise performs better than the no-noise version in terms of MAE, this observation in the PAD maps confirms the inclusion of noise does not pose a hindrance or concern in the proposed methodology. 

\nw{The proposed model's performance on the sex-stratified CC359 subsets as reported in the results section shows that sex is an important confounder to be considered in brain age prediction studies. The aging process and related structural changes in GM, WM, and CSF vary across different sexes  \citep{gur1999sex,wang2019effects} and is reflected in the difference in MAEs on the two sex-specific test sets. To minimize any bias caused by aging differences across males and females, the proposed methodology utilized nearly balanced datasets (for training as well as testing) to ensure equal representation of both sexes. However, future work can include attempts to model voxel-level brain age for sex-specific populations to study the influence of sex in more detail. }

The proposed model produces voxel-level PAD maps, which are compared to the heatmaps obtained from traditional interpretability methods. An important feature of the proposed approach that contributes towards ensuring that the proposed model is learning correct features from the input image is the addition of the brain tissue segmentation task as one of the outputs in the architecture. Owing to the multitasking design, the model re-uses the features for the segmentation as well as brain age prediction task. The segmentation performance of the proposed model reached a dice score of \nw{0.89 and 0.83 on the Cam-CAN and CC359 datasets respectively,} indicating substantial overlap between predictions and ground truth segmentations. A considerable performance on the segmentation task goes to show that the model learns the structural intricacies within the brain volume \nw{such as thickness and shape among other features} which are then repurposed for the voxel-level brain age prediction task. This confirms that \nw{structural features like changes in GM, WM, and CSF volume, thickness, shape etc. drive model predictions rather than extraneous noise in the background. The inclusion of background noise or any irrelevant features being learned by the proposed model would impact the segmentation performance negatively and be reflected in the dice score, however, that is not the case as the model achieves a considerable dice score on both the internal and external test sets.} 

Contrary to the heatmaps obtained from traditional interpretability methods which are based on gradients with respect to an input (Grad-CAM, SmoothGrad), the voxel-level PAD maps reflect differences in the prediction from the chronological age in years, making them quantitative and comparable across samples. The occlusion sensitivity maps come close to voxel-PAD maps, however, they are generated by occluding a single region at a time and evaluating its impact on the global age prediction. It is vital to acknowledge that in most machine learning models, multiple regions, which might not adhere to square or cuboid structures, collectively influence final predictions, thus, assessing these regions in isolation is informative, but does not provide the most accurate insight into the collective contributions to brain age predictions. PAD maps, on the other hand, utilize structural features within the brain region and reflect on voxel-level brain age instead of global brain age, and the results show that the spatial differences in the aging process observed make clinical sense when compared against the structural changes in corresponding T1-weighted images \citep{gianchandani2023}.

The regional analysis of the PAD maps corresponding to presumed healthy subjects shows PAD values in the narrow range of -1.96 years to 1.87 years \nw{(Figure \ref{mni_atlas_Cam-CAN})}, \textit{i.e.}, making most regions (except three) appear slightly younger than the expected chronological age, however, the difference is minimal and can be \nw{possibly} accounted for by the modeling error. The \nw{average regional PAD} are closely aligned near 0 (brain age $=$ chronological age), which is the ideal and theoretical scenario, however, does not account for the spatial variations observed in the brain ages across different regions and different samples. \nw{It must also be noted that the S.D. (Table \ref{tab:regionalpad} - column Average regional PAD) is considerably large for healthy subjects which points to a significant subject-to-subject variability.} However, it must be kept in mind that this analysis pertains to a population level encompassing subjects with a diverse age range and unique trajectories of brain aging, \nw{all of which might not be reflective in population-level regional averages. Keeping in mind that the aging of the brain is unique to each individual, a subject-level analysis will provide better estimations of regional aging trajectories for the individual.} 

For the regional analysis, subjects with age $\leq$ 70 are filtered for the test set. There are two reasons for doing so: (i) The proposed model is trained on subjects up to 88 years of age and to maintain the reliability of predictions, a deliberate choice was made to refrain from evaluating the model on subjects exceeding 88 years of age. The predictions in the peripheral regions of the \nw{in-domain Cam-CAN test set} (ages 70 and above as shown in Figure \ref{scatter_plot}) are often observed to exhibit a  bias, leading to under-prediction or younger-looking brains for older age ranges. While the bias is addressed through a dedicated correction process as explained in Section \ref{subsec:biascorrection}, it is important to note that the methodology used for this bias correction is built upon data from healthy subjects. It is tailored to the patterns observed in the evaluation of healthy subjects. It would be unfair to assume that, for diseased subjects, the same bias correction methodology would suffice to mitigate the bias observed.
(ii) Based on the bias-correction methodology, a different correcting factor is used for different age ranges and theoretically, if diseased subjects are expected to have an increased brain age relative to the corresponding chronological age, it would be unfair to use the correcting factor based on the chronological age as the bias observed would be relative to an older age (compared to the chronological age). Hence, to ensure that bias correction does not fail significantly, and helps with mitigating the bias to a reasonable extent, this precautionary filtering is performed to remove subjects with age $\geq$ 71 years. Nonetheless, since most neurological disorders are observed in an older population, bias correction becomes imperative for the AD and dementia test sets for the regional analysis to help account for the bias, even though it might not mitigate the bias entirely. \nw{For consistency of results, precautionary filtering and bias-correction were also performed for the healthy test set for regional analysis.}

Another important consideration when analyzing the regional PAD values in Table \ref{tab:regionalpad} is that in the case of a healthy population, the age range of subjects is wide enough \nw{(18-70 years)} such that the small bias observed is in both directions as over-predictions and as well as under-predictions. Hence, at the population level, the over and under-predictions tend to cancel each other's effect to an extent. However, this might not be the case for diseased subjects as most subjects in the test set are above the average training set age and hence, bias is only observed in the form of under-predictions (\textit{i.e.} negative PAD). As mentioned previously, bias-correction does not account for 100\% of the bias in diseased subjects coupled with the fact that only under-predictions are observed, the results of the PAD values in Table \ref{tab:regionalpad} and \Cref{mni_atlas_alzheimers,mni_atlas_Dementia} might still reflect a small degree of bias and be more negative than the actual values.

The regional PAD values, MNI atlases, and PAD maps corresponding to individual subjects were reviewed by a radiologist (JO) and some notable observations were made: 

1. It can be observed that in subjects with dementia, ventricles tend to show an accelerated brain age as compared to the rest of the brain regions (refer to adjusted PAD maps in Figure \ref{oasis_diseased}). It is unclear whether this increased aging of the ventricles is mostly related to an increase in ventricle size, which is usually a sequelae of generalized brain parenchymal volume loss, or due to differences in CSF composition. Both these explanations seem plausible: large ventricle size is associated with the presence of neurodegenerative disorders, and even in healthy subjects, increased ventricle volume seems to indicate a greater risk of developing dementia in the future \citep{carmichael2007ventricular}. Furthermore, cellular CSF composition is altered in subjects with neurodegenerative diseases, with a shift from central memory to effector T cells \citep{busse2021Dementia}. Such changes do not affect MR image signal intensity in any noticeable way upon visual inspection by radiologists, but there may be subtle signal changes that may have been detected by the proposed model.\\
2. In AD subjects, PAD was particularly high in the Caudate nuclei \nw{with an average of $\sim$ 4.29 years} (Figure \ref{mni_atlas_alzheimers}). Previous studies have found lower Caudate nuclei volumes in AD compared to healthy control subjects \citep{madsen20103d}. \nw{Considering the possible association between} lower Caudate volume and advanced brain age, these prior findings are in line with the results of the current study. \nw{It is important to emphasize that this is a speculative interpretation of the results presented in this work and should not be translated to clinical applications without further analysis.} Increased brain age (2.8 years) was also observed in the Temporal Lobe with a high regional standard deviation indicating a great degree of variation within the region, which is often an important region associated with AD.\\
3. In the group of dementia subjects, brain age was particularly advanced in the posterior brain regions, \textit{i.e.}, the Occipital and posterior Parietal lobes, and the Cerebellum (Figure \ref{mni_atlas_Dementia}). Atrophy predominantly affecting the posterior brain parenchyma is uncommon in dementia patients. It can sometimes be seen in AD patients \citep{crutch2012posterior} and is a hallmark feature of Lewy body dementia, a rare neurodegenerative disease \citep{silva2023characteristics}. \nw{The exact underlying dementia etiologies are not known in the dementia subgroup of this study; there is a possibility that some of these subjects did suffer from Lewy body dementia or posterior predominant AD.} While previous studies mainly focused on brain parenchymal volume, the proposed model predicts brain age using a multidimensional approach. It is possible that characteristics other than volume, for example, changes in brain signal intensity or structure, occur in subjects with dementia that do not affect volume and are, therefore, not well known yet. \nw{This, however, is speculative since we do not have exact clinical diagnoses and therefore can neither confirm nor refute this hypothesis.}\\
4. \nw{In the group of dementia subjects, negative PAD was observed in certain regions like the Cuadate, Frontal Lobe, Insula, and Putamen. Since the exact underlying dementia etiologies are not known in this group, the negative values could be a grouped effect of multiple different types of dementia observed (or not observed) in the subjects of this group. This group showed increased brain age in the posterior regions of the brain which points to the possible presence of subjects with posterior predominant dementia, whereas a negative PAD for the Frontal Lobe indicates the absence of subjects with frontal predominant dementia. Insular atrophy or degeneration is also observed in the presence of frontotemporal dementia \citep{seeley2010anterior} and could be the driving factor for a negative PAD. Putamen is dominantly associated with the presence of Parkinson's disease \citep{kinoshita2022putamen}, however, it averaged at -0.31 years which is a rather small deviation from zero and could simply be an impact of modeling error. It must also be noted that a high standard deviation (Table \ref{tab:regionalpad}: Avg Regional PAD column) is observed for the dementia test set as compared to others that indicates subject-to-subject variation, which is plausible as this test set includes subjects with different dementia types and hence, differences in regional brain aging trajectories.} 

The findings from the the proposed brain age prediction model are partially consistent with the known biomarkers of aging in subjects with dementia and more specifically, AD. Some new potential biomarkers like increased brain age in posterior regions of the brain have been identified by the proposed model, and require further validation. 

It is crucial to emphasize that though it is important to understand regional aging patterns for older subjects \textit{i.e.}, where disorders are observed and are often progressed to a stage where the subject exhibits noticeable symptoms and is already a part of the research study collecting data; another important aim of this research is to predict early onset of neurological disorders before the subjects start exhibiting symptoms and apparent cognitive decline. Therefore, evaluation on healthy subjects is an important part as it can unveil potential indicators of early onset of neurological disorders. A future direction to validate the proposed model would be to evaluate the model on longitudinal data which includes subjects transitioning from an initial presumed healthy stage to some form of underlying neurological disorder.

\section{Conclusion}
\label{sec:conclusion}
In this study, previous analysis of a voxel-level brain age prediction model is extended as a proof-of-concept. Through the experiments, the choice of a multitask architecture is validated and it is shown that using a voxel-level approach can be a way of achieving improved interpretability and a better understanding of regional aging trajectories. Evaluation of the model on healthy subjects as well as ones with dementia and specifically, AD revealed consistent findings on regional brain aging as other aging studies and also revealed new indicators that can be potential biomarkers of the presence of dementia. Through this research, the transition of brain age prediction models towards voxel-level predictions is shown as a way to enhance the understanding of the degenerating brain while demonstrating an improvement with respect to existing implementations.


\acks{NG is supported by the Natural Sciences and Engineering Research Council (NSERC) BRAIN CREATE award and the Alberta Innovates Graduate Student Scholarship. RS thanks the NSERC (RGPIN/2021-02867) for ongoing operating support for this project. RS also thanks the Hotchkiss Brain Institute for financial support. MEM acknowledges support from startup funding at the University of Calgary and the NSERC Discovery Grant (RGPIN-03552) and Early Career Researcher Supplement (DGECR-00124). Data collection and sharing for this project was partly funded by the Alzheimer's Disease Neuroimaging Initiative (ADNI) (National Institutes of Health Grant U01 AG024904) and DOD ADNI (Department of Defense award number W81XWH-12-2-0012).}

%
\ethics{The work follows appropriate ethical standards in conducting research and writing the manuscript. All data used in this study was obtained from publicly available datasets and has been handled following the terms provided by the data sources. Data anonymity has been maintained and all data sources have been properly cited complying with ethical and privacy regulations.}

\coi{The authors have no competing interests to declare.}

\bibliography{references}


\end{document}